\documentclass{article} 
\usepackage{nips14submit_e,times}
\usepackage{hyperref}
\usepackage{url}
\usepackage{natbib}
\usepackage{amsmath}
\usepackage{amsthm}
\usepackage{amssymb}
\usepackage{graphicx}
\usepackage{xspace}
\usepackage{tabularx}

\newcommand{\obs}{\text{obs}}
\newcommand{\mis}{\text{mis}}

\newcommand{\qt}[1]{\left<#1\right>}

\newcommand{\hl}{HL}

\newcommand{\bmx}[0]{\begin{bmatrix}}
\newcommand{\emx}[0]{\end{bmatrix}}

\newcommand{\vect}[1]{\mathbf{#1}}

\newcommand{\matr}[1]{\mathbf{#1}}
\newcommand{\var}[0]{\operatorname{Var}}

\newcommand{\vb}[0]{\vect{b}}
\newcommand{\vc}[0]{\vect{c}}
\newcommand{\vh}[0]{\vect{h}}
\newcommand{\vv}[0]{\vect{v}}
\newcommand{\vx}[0]{\vect{x}}

\newcommand{\vm}[0]{\vect{m}}

\newcommand{\mW}[0]{\matr{W}}

\newcommand{\mV}[0]{\matr{V}}

\newcommand{\sigmoid}{\sigma}
\newcommand{\E}[0]{\mathbb{E}}

\newcommand{\dd}[1]{\ensuremath{\mbox{d}#1}}

\newcommand{\NADEk}{NADE-$k$}

\title{Iterative Neural Autoregressive Distribution Estimator (\NADEk{})}

\author{
Tapani Raiko\\
Aalto University
\And
Li Yao\\
Universit\'{e} de Montr\'{e}al
\And
KyungHyun Cho\\
Universit\'{e} de Montr\'{e}al
\And
Yoshua Bengio\\
Universit\'{e} de Montr\'{e}al, \\
CIFAR Senior Fellow
}

\nipsfinalcopy 

\begin{document}

\maketitle

\begin{abstract}
Training of the neural autoregressive density estimator (NADE) can be viewed as
doing one step of probabilistic inference on missing values in data.
We propose a new model that extends this inference scheme to multiple steps,
arguing that it is easier to learn to improve a reconstruction in $k$ steps
rather than to learn to reconstruct in a single inference step.  The proposed
model is an unsupervised building block for deep learning that combines the
desirable properties of NADE and multi-prediction training: (1) Its test
likelihood can be computed analytically, (2) it is easy to generate
independent samples from it, and (3) it uses an inference engine that is
a superset of variational inference for Boltzmann machines. 
The proposed NADE-k is competitive with the state-of-the-art in density estimation 
on the two datasets tested. 
\end{abstract}

\section{Introduction}

Traditional building blocks for deep learning have some unsatisfactory
properties. 
Boltzmann machines
are, for instance, difficult to train due to the intractability of computing the
statistics of the model distribution, which leads to the potentially high-variance MCMC estimators
during training (if there are many well-separated modes~\citep{Bengio-et-al-ICML2013})
and the computationally intractable objective function.
Autoencoders have a simpler objective function (e.g., denoising reconstruction
error~\citep{Vincent-JMLR-2010-small}), which can be used for model selection
but not for the important choice of the corruption function.  On the other
hand, this paper follows up on the Neural Autoregressive Distribution
Estimator~\citep[NADE,][]{larochelle2011neural}, which specializes previous
neural auto-regressive density estimators~\citep{Bengio+Bengio-NIPS99} and was
recently extended~\citep{Benigno-et-al-icml-2014} to deeper architectures. It
is appealing because both the training criterion (just log-likelihood) and its
gradient can be computed tractably and used for model selection,
and the model can be trained by stochastic gradient descent with
backpropagation. However, it has been observed that the performance of NADE has
still room for improvement.

The idea of using missing value imputation as a training criterion has appeared
in three recent papers.
This approach
can be seen either as training an energy-based model to impute missing values
well~\citep{brakel2013training}, as training a generative probabilistic model
to maximize a generalized pseudo-log-likelihood~\citep{goodfellow2013multi}, or
as training a denoising autoencoder with a masking corruption
function~\citep{Benigno-et-al-icml-2014}.  Recent work on generative stochastic
networks (GSNs), which include denoising auto-encoders as special cases,
justifies dependency networks~\citep{HeckermanD2000} as well as generalized
pseudo-log-likelihood~\citep{goodfellow2013multi}, but have the disadvantage
that sampling from the trained ``stochastic fill-in'' model requires a Markov
chain (repeatedly resampling some subset of the values given the others).
In all these cases, learning progresses by back-propagating the imputation
(reconstruction) error through inference steps of the model. This allows the
model to better cope with a potentially imperfect inference algorithm.  This
learning-to-cope was introduced recently in 2011 by
\citet{stoyanov2011empirical} and \citet{domke2011parameter}.

The NADE model involves an ordering over the components of the data vector. The
core of the model is the reconstruction of the next component given all the
previous ones.  In this paper we reinterpret the reconstruction procedure as a
single iteration in a variational inference algorithm, and we propose a version
where we use $k$ iterations instead, inspired by
\citep{goodfellow2013multi,brakel2013training}.  We evaluate the proposed model
on two datasets and show that it outperforms the original
NADE~\citep{larochelle2011neural} as well as NADE trained with the
order-agnostic training algorithm~\citep{Benigno-et-al-icml-2014}.

\section{Proposed Method: \NADEk{}}
\label{seq:proposed_method}

We propose a probabilistic model called \NADEk{} for $D$-dimensional binary
data vectors $\vx$. We start by defining $p_{\boldsymbol{\theta}}$ for imputing
missing values using a fully factorial conditional distribution:
\begin{align}
  p_{\boldsymbol{\theta}}(\vx_\mis \mid \vx_\obs) = \prod_{i \in \mis} p_{\boldsymbol{\theta}}(x_i \mid \vx_\obs),
  \label{eq:imputation}
\end{align}
where the subscripts $\mis$ and $\obs$ denote missing and observed components
of $\vx$. From the conditional distribution $p_{\boldsymbol{\theta}}$ we
compute the joint probability distribution over $\vx$ given an ordering $o$ (a
permutation of the integers from $1$ to $D$) by
\begin{align}
    p_{{\boldsymbol{\theta}}}(\vx \mid o) = \prod_{d=1}^D p_{\boldsymbol{\theta}}(x_{o_d} \mid \vx_{o_{<d}}),
  \label{eq:joint_given_ordering}
\end{align}
where $o_{<d}$ stands for indices $o_1\dots o_{d-1}$.

The model is trained to minimize the negative log-likelihood averaged over all
possible orderings $o$
\begin{align}
  \mathcal{L}({\boldsymbol{\theta}})=\E_{o \in D!} \left[ 
      \E_{\vx \in \text{data}} \left[- \log p_{\boldsymbol{\theta}} (\vx \mid
  o)\right]\right].
  \label{eq:criterion_expected}
\end{align}
using an unbiased, stochastic estimator of $\mathcal{L}({\boldsymbol{\theta}})$ 
\begin{equation}
  \hat{\mathcal{L}}({\boldsymbol{\theta}}) = - \frac{D}{D-d+1} \log p_{\boldsymbol{\theta}} (\vx_{o_{\geq d}} \mid \vx_{o_{<d}})
  \label{eq:criterion_sampled}
\end{equation}
by drawing $o$ uniformly from all $D!$ possible orderings and $d$ uniformly
from $1\dots D$~\citep{Benigno-et-al-icml-2014}. Note that while the model
definition in Eq.\ \eqref{eq:joint_given_ordering} is sequential in nature, the
training criterion \eqref{eq:criterion_sampled} involves reconstruction of all
the missing values in parallel. In this way, training does not involve picking
or following specific orders of indices.

In this paper, we define the conditional model
$p_{\boldsymbol{\theta}}(\vx_\mis \mid \vx_\obs)$ using a deep feedforward
neural network with $nk$ layers, where we use $n$ weight matrices $k$ times.
This can also be interpreted as running $k$ successive inference steps with an
$n$-layer neural network. 

\begin{figure}[t]
\centering
\includegraphics[width=0.27\textwidth]{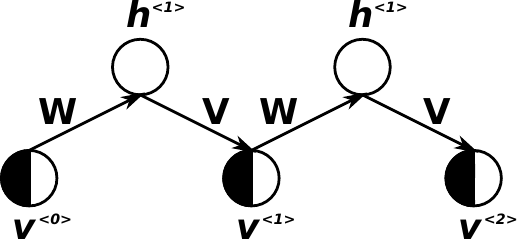}
\hfill
\includegraphics[width=0.27\textwidth]{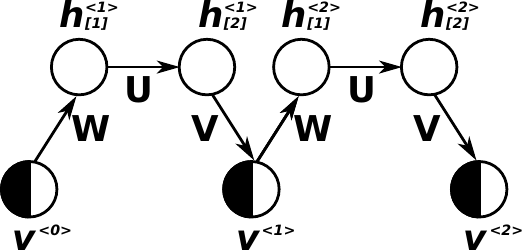}
\hfill
\includegraphics[width=0.38\textwidth]{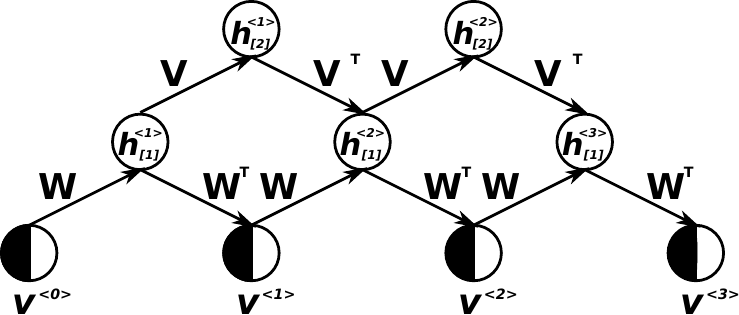}
\caption{
    The choice of a structure for \NADEk{} is very flexible. The dark filled halves
    indicate that a part of the input is observed and fixed to the observed
    values during the iterations. Left: Basic structure
    corresponding to Equations (\ref{eq:iter_h}--\ref{eq:iter_v}) with $n=2$ and
    $k=2$. Middle: Depth added as in NADE by \citet{Benigno-et-al-icml-2014} with $n=3$ and
    $k=2$. Right: Depth added as in Multi-Prediction Deep Boltzmann Machine by
    \citet{goodfellow2013multi} with $n=2$ and $k=3$. The first two structures are 
    used in the experiments.
}
\label{fig:structures}
\end{figure}

The input to the network is 
\begin{align}
    \vv^{\qt{0}} &= \vm \odot \E_{\vx \in
        \text{data}} \left[ \vx \right] + (\vect{1}-\vm) \odot \vx 
    \label{eq:v0}
\end{align}
where $\vm$ is a binary mask vector indicating missing components with 1, and
$\odot$ is an element-wise multiplication.  $\E_{\vx \in \text{data}} \left[
\vx \right]$ is an empirical mean of the observations.  For simplicity, we give
equations for a simple structure with $n=2$.  See Fig.~\ref{fig:structures}
(left) for the illustration of this simple structure.

\begin{figure}[t]
\begin{minipage}{0.49\textwidth}
    \centering
    \includegraphics[width=\textwidth]{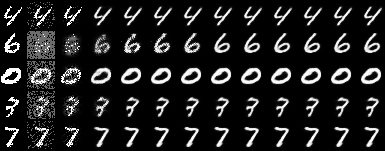} 
\end{minipage}
\begin{minipage}{0.49\textwidth}
    \centering
    \includegraphics[width=\textwidth]{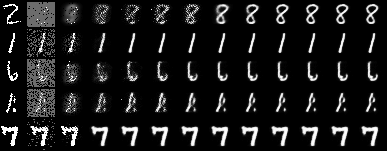}
\end{minipage}
\caption{The inner working mechanism of \NADEk{}. The left most column shows the
    data vectors $\vx$, the second column shows their masked
    version and the subsequent columns show the reconstructions $\vv^{\qt{0}}\dots \vv^{\qt{10}}$ (See Eq.~\eqref{eq:iter_v}).
}
\label{fig:convergence}
\end{figure}

In this case, the activations of the 
layers at the $t$-th step are
\begin{align}
  \label{eq:iter_h}
  \vh^{\qt{t}} &= \phi (\mW \vv^{\qt{t-1}} + \vc) \\
  \label{eq:iter_v}
    \vv^{\qt{t}} &= 
    \vm \odot \sigmoid (\mV \vh^{\qt{t}} + \vb) + (\vect{1}-\vm) \odot \vx
\end{align}
where $\phi$ is an element-wise nonlinearity, $\sigma$ is a logistic sigmoid
function, and the iteration index $t$ runs from $1$ to $k$. The conditional
probabilities of the variables (see Eq.~\eqref{eq:imputation}) are read from
the output $\vv^{\qt{k}}$ as
\begin{align}
  p_{\boldsymbol{\theta}}(x_i = 1\mid \vx_\obs) &= v_i^{\qt{k}}.
  \label{eq:v_to_prob}
\end{align}
Fig.~\ref{fig:convergence} shows examples of how $\vv^{\qt{t}}$ evolves over
iterations, with the trained model.

The parameters ${\boldsymbol{\theta}} = \left\{\mW, \mV, \vc, \vb\right\}$ can
be learned by stochastic gradient descent to minimize 
$-\mathcal{L}({\boldsymbol{\theta}})$ in Eq.~\eqref{eq:criterion_expected}, or
its stochastic approximation $-\hat{\mathcal{L}}({\boldsymbol{\theta}})$ in
Eq.~\eqref{eq:criterion_sampled}, with the stochastic gradient computed by
back-propagation. 

Once the parameters ${\boldsymbol{\theta}}$ are learned, we can define a
mixture model by using a uniform probability over a set of orderings $O$. We
can compute the probability of a given vector $\vx$ as a mixture model
\begin{align}
  \label{eq:ensemble}
  p_\text{mixt}(\vx\mid {\boldsymbol{\theta}}, O) = \frac{1}{\left|O\right|} \sum_{o \in O} p_{\boldsymbol{\theta}}(\vx\mid o)
\end{align}
with Eq.~\eqref{eq:joint_given_ordering}. We can draw independent samples from
the mixture by first drawing an ordering $o$ and then sequentially drawing each
variable using $x_{o_d}\sim p_{\boldsymbol{\theta}}(x_{o_d}\mid \vx_{o_{<d}})$.
Furthermore, we can draw samples from the conditional $p(\vx_{\mis} \mid
\vx_\obs)$ easily by considering only orderings where the observed indices
appear before the missing ones. 

{\bf Pretraining}
It is well known that training deep networks is difficult without pretraining,
and in our experiments, we train networks up to $kn=7\times 3=21$ layers.  When
pretraining, we train the model to produce good reconstructions $\vv^{\qt{t}}$
at each step $t=1\dots k$. More formally, in the pretraining phase, we replace
Equations (\ref{eq:criterion_sampled}) and (\ref{eq:v_to_prob}) by
\begin{align}
  \label{eq:pretraining}
  &\hat{\mathcal{L}}_\text{pre}({\boldsymbol{\theta}}) 
  =  - \frac{D}{D-d+1} \frac{1}{k} \sum_{t=1}^k \log \prod_{i \in o_{\geq d}} p^{\qt{t}}_{\boldsymbol{\theta}}(x_i \mid \vx_{o_{<d}})\\
  &p^{\qt{t}}_{\boldsymbol{\theta}}(x_i = 1\mid \vx_\obs) = v_i^{\qt{t}}.
\end{align}

\subsection{Related Methods and Approaches}

{\bf Order-agnostic NADE}
The proposed method follows closely the order-agnostic version of
NADE~\citep{Benigno-et-al-icml-2014}, which may be considered as the special
case of \NADEk{} with $k=1$. On the other hand, \NADEk{} can be seen as a deep
NADE with some specific weight sharing (matrices $\mW$ and $\mV$ are reused for
different depths) and gating in the activations of some layers (See Equation
\eqref{eq:iter_v}). 

Additionally, \citet{Benigno-et-al-icml-2014} found it crucial to give the mask
$\vm$ as an auxiliary input to the network, and initialized missing values to
zero instead of the empirical mean (See Eq.~\eqref{eq:v0}). Due to these
differences, we call their approach NADE-mask.  One should note that NADE-mask
has more parameters due to using the mask as a separate input to the network,
whereas \NADEk{} is roughly $k$ times more expensive to compute.

{\bf Probabilistic Inference}
Let us consider the task of missing value imputation in a probabilistic latent
variable model. We get the conditional probability of interest by marginalizing
out the latent variables from the posterior distribution:
\begin{align}
    p(\vx_\mis \mid \vx_\obs) = \int_\vh p(\vh, \vx_\mis \mid \vx_\obs) \dd{\vh}.
\end{align}

Accessing the joint distribution $p(\vh, \vx_\mis \mid \vx_\obs)$ directly is
often harder than alternatively updating $\vh$ and $\vx_\mis$ based on
the conditional distributions $p(\vh \mid \vx_\mis, \vx_\obs)$ and $p(\vx_\mis
\mid \vh)$.\footnote{
    We make a typical assumption that observations are mutually independent
    given the latent variables.
} Variational inference is one of the representative examples that exploit this.

In variational inference, a factorial distribution
$q(\vh,\vx_\mis)=q(\vh)q(\vx_\mis)$ is iteratively fitted to $p(\vh, \vx_\mis
\mid \vx_\obs)$ such that the KL-divergence between $q$ and $p$
\begin{align}
  \text{KL}[q(\vh, \vx_\mis) || p(\vh,\vx_\mis \mid \vx_\obs)] = 
  - \int_{\vh,\vx_\mis} q(\vh,\vx_\mis) \log \left[\frac{p(\vh, \vx_\mis \mid
  \vx_\obs)}{q(\vh,\vx_\mis)}\right] \dd{\vh} \dd{\vx_\mis}
\end{align}
is minimized. The algorithm alternates between updating $q(\vh)$ and
$q(\vx_\mis)$, while considering the other one fixed.

As an example, let us consider a restricted Boltzmann machine (RBM) defined by
\begin{align}
  p(\vv,\vh) \propto \exp(\vb^\top \vv + \vc^\top \vh + \vh^\top \mW \vv).
\end{align}
We can fit an approximate posterior distribution parameterized as
$q(v_i=1)=\bar{v}_i$ and $q(h_j=1)=\bar{h}_j$ to the true posterior
distribution by iteratively computing
\begin{align}
 \bar{\vh} &\leftarrow \sigma(\mW \bar{\vv} + \vc) \\
 \bar{\vv} &\leftarrow \vm \odot \sigmoid (\mW^\top \vh + \vb) + (\vect{1}-\vm) \odot \vv.
\end{align}
We notice the similarity to Eqs.~\eqref{eq:iter_h}--\eqref{eq:iter_v}: If we
assume $\phi=\sigmoid$ and $\mV=\mW^\top$, the inference in the \NADEk{} is
equivalent to performing $k$ iterations of variational inference on an RBM for
the missing values~\citep{Peterson1987}. We can also get variational inference on
a deep Boltzmann machine (DBM) using the structure in Fig.~\ref{fig:structures}
(right).


{\bf Multi-Prediction Deep Boltzmann Machine}
\citet{goodfellow2013multi} and \citet{brakel2013training} use backpropagation
through variational inference steps to train a deep Boltzmann machine. This is
very similar to our work, except that they approach the problem from the view
of maximizing the generalized pseudo-likelihood~\citep{Huang2002}.
Also, the deep Boltzmann machine lacks the tractable probabilistic
interpretation similar to \NADEk{} (See Eq.~\eqref{eq:joint_given_ordering})
that would allow to compute a probability or to generate independent samples
without resorting to a Markov chain. Also, our approach is somewhat more
flexible in the choice of model structures, as can be seen in
Fig.~\ref{fig:structures}.
For instance, in the proposed NADE-$k$, encoding and decoding weights do not
have to be shared and any type of nonlinear activations, other than a logistic
sigmoid function, can be used.


{\bf Product and Mixture of Experts}
One could ask what would happen if we would define an ensemble likelihood along
the line of the training criterion in Eq.~\eqref{eq:criterion_expected}. That
is,
\begin{align}
  -\log p_\text{prod}(\vx \mid {\boldsymbol{\theta}}) \propto \E_{o \in D!}
  \left[-
  \log p(\vx \mid {\boldsymbol{\theta}}, o)\right].
\end{align}
Maximizing this ensemble likelihood directly will correspond to training a
product-of-experts model~\citep{Hinton-PoE-2000}. 
However, this requires us to evaluate the intractable normalization constant
during training as well as in the inference, making the model not tractable
anymore.

On the other hand, we may consider using the log-probability of a sample under
the mixture-of-experts model as the training criterion
\begin{align}
  - \log p_\text{mixt}(\vx \mid {\boldsymbol{\theta}}) = - \log \E_{o \in
  D!}\left[
  p(\vx \mid {\boldsymbol{\theta}}, o)\right].
\end{align}
This criterion resembles clustering, where individual models may specialize in
only a fraction of the data. In this case, however, the simple estimator such
as in Eq.~\eqref{eq:criterion_sampled} would not be available.

\section{Experiments}
We study the proposed model with two datasets: binarized MNIST handwritten
digits and Caltech 101 silhouettes.

We train \NADEk{} with one or two hidden layers ($n=2$ and $n=3$, see
Fig.~\ref{fig:structures}, left and middle) with a hyperbolic tangent as the
activation function $\phi(\cdot)$. We use stochastic gradient descent on
the training set with a minibatch size fixed to $100$. We use
AdaDelta~\citep{Zeiler-2012} to adaptively choose a learning rate for each
parameter update on-the-fly.  We use the validation set for early-stopping and
to select the hyperparameters.  With the best model on the validation set, we
report the log-probability computed on the test set. 
We have made our implementation available\footnote{git@github.com:yaoli/nade\_k.git}.

\subsection{MNIST}

We closely followed the procedure used by \citet{Benigno-et-al-icml-2014},
including the split of the dataset into 50,000 training samples, 10,000
validation samples and 10,000 test samples. We used the same version where the
data has been binarized by sampling.

We used a fixed width of 500 units per hidden layer.  The number of steps $k$
was selected among $\{ 1, 2, 4, 5, 7\}$. According to our preliminary
experiments, we found that no separate regularization was needed when using a
single hidden layer, but in case of two hidden layers, we used weight decay
with the regularization constant in the interval $\left[ e^{-5}, e^{-2}
\right]$.  Each model was pretrained for 1000 epochs and fine-tuned for 1000
epochs in the case of one hidden layer and 2000 epochs in the case of two.

For both \NADEk{} with one and two hidden layers, the validation performance
was best with $k=5$. The regularization constant was chosen to be 0.00122 for
the two-hidden-layer model.

{\bf Results}
We report in Table~\ref{tab:mnist_benchmark} the mean of the test
log-probabilities averaged over randomly selected orderings. We also show the
experimental results by others from \citep{Benigno-et-al-icml-2014,Gregor-et-al-ICML2014}. 
We denote
the model proposed in \citep{Benigno-et-al-icml-2014} as a \textit{NADE-mask}.

\begin{table}[t]
\centering
\begin{tabular}{|l | c || l | r |}
\hline
\centering Model  & Log-Prob. & Model & Log-Prob. \\
\hline
\hline
NADE 1\hl (fixed order)      & -88.86 & RBM (500h, CD-25)           & $\approx$ -86.34\\
NADE 1\hl       & -99.37  & DBN (500h+2000h)            & $\approx$ {-84.55}\\
NADE 2\hl       & -95.33  & DARN (500h) & $\approx$ -84.71 \\
NADE-mask 1\hl                  & -92.17  & DARN (500h, adaNoise) & $\approx$ {\bf -84.13} \\
\cline{3-4}
NADE-mask 2\hl                  & -89.17  & NADE-5 1\hl                  & -90.02\\
NADE-mask 4\hl                  & -89.60  & NADE-5 2\hl                  & -87.14\\
EoNADE-mask 1\hl (128 Ords)& -87.71  & EoNADE-5 1\hl (128 Ords)& -86.23 \\
EoNADE-mask 2\hl (128 Ords)& -85.10  & EoNADE-5 2\hl (128 Ords)& {\bf -84.68} \\
\hline
\end{tabular}
\caption{
    Results obtained on MNIST using various models and number of hidden
    layers (1\hl{} or 2\hl). ``Ords'' is short for ``orderings''. These are the average
    log-probabilities of the test set. EoNADE refers to the ensemble probability
    (See Eq.~\eqref{eq:ensemble}). From here on, in all figures and tables we
    use ``\hl'' to denote the number of hidden layers and ``h'' for the number of
    hidden units.
}
\label{tab:mnist_benchmark}
\end{table}

From Table~\ref{tab:mnist_benchmark}, it is clear that \NADEk{} outperforms the
corresponding NADE-mask 
both with the individual orderings and ensembles over orderings using both 1 or
2 hidden layers.  \NADEk{} with two hidden layers achieved the generative
performance comparable to that of the deep belief network (DBN) with two hidden
layers.

\begin{figure}[t]
\begin{minipage}{0.49\textwidth}
\centering
\includegraphics[width=\textwidth]{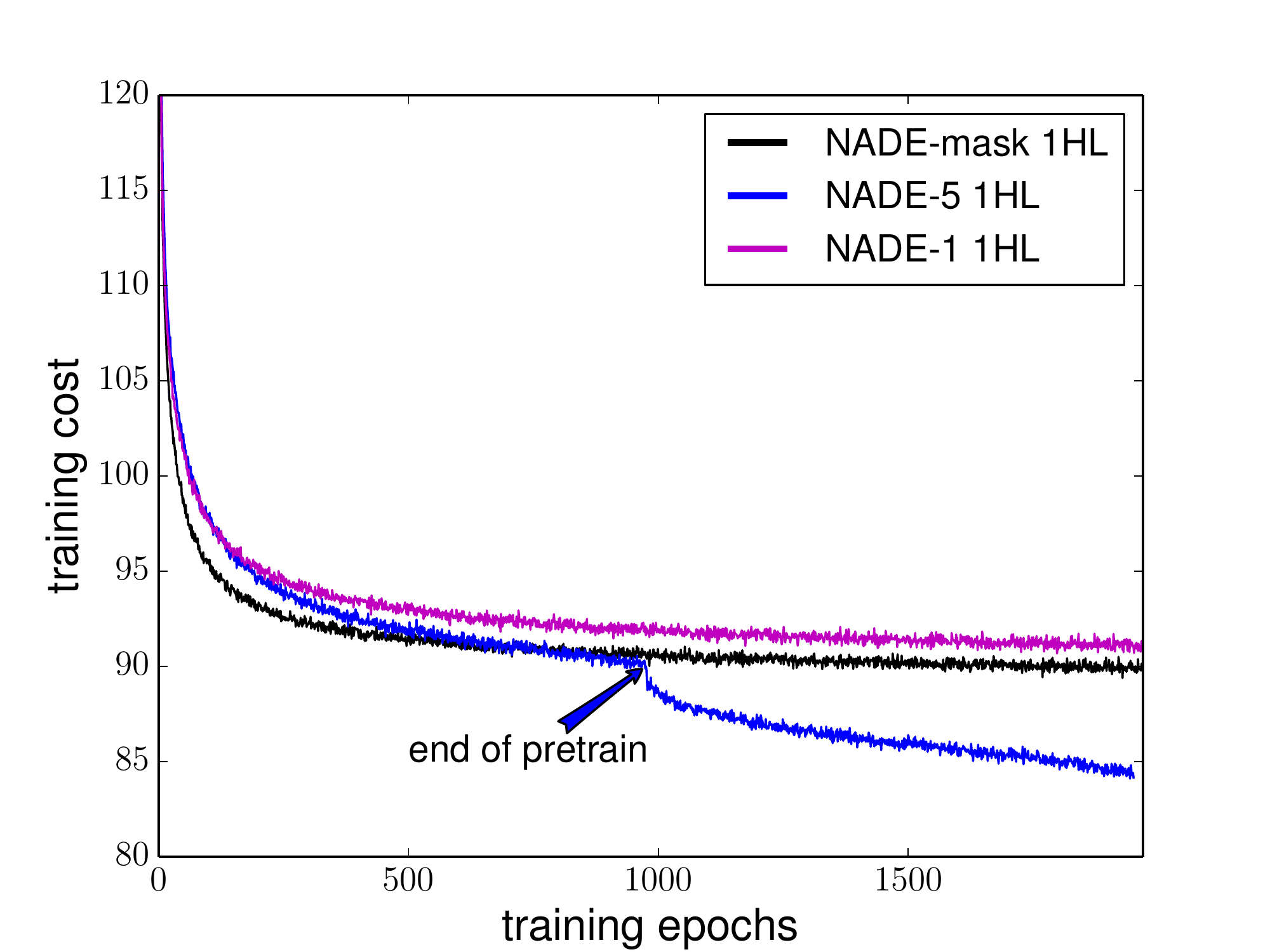} 
(a)
\end{minipage}
\begin{minipage}{0.49\textwidth}
\centering
\includegraphics[width=\textwidth]{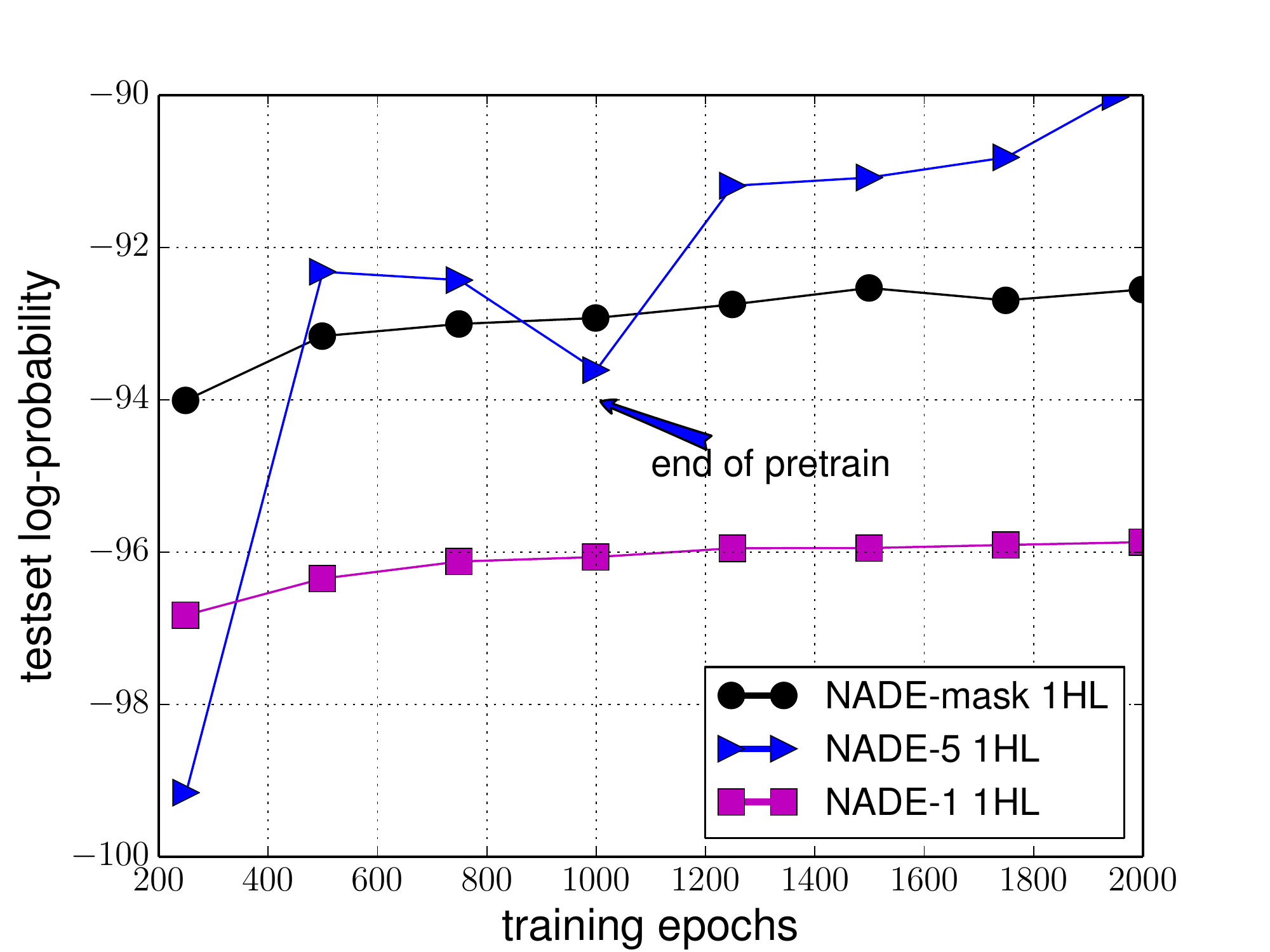}
(b)
\end{minipage}
\caption{\NADEk{} with k steps of variational inference helps to reduce  the 
training cost (a) and to generalize better (b). NADE-mask performs better than NADE-1 
without masks both in training and test.}
\label{fig:learning_dynamics}
\end{figure}

Fig.~\ref{fig:learning_dynamics} shows training curves for some of the models.
We can see that the NADE-$1$ does not perform as well as NADE-mask. This
confirms that in the case of $k=1$, the auxiliary mask input is indeed useful.
Also, we can note that the performance of NADE-5 is still improving at the end
of the preallocated 2000 epochs, further suggesting that it may be possible to
obtain a better performance simply by training longer.

\begin{figure}[t]
\begin{minipage}{0.49\textwidth}
\centering
\includegraphics[width=\textwidth]{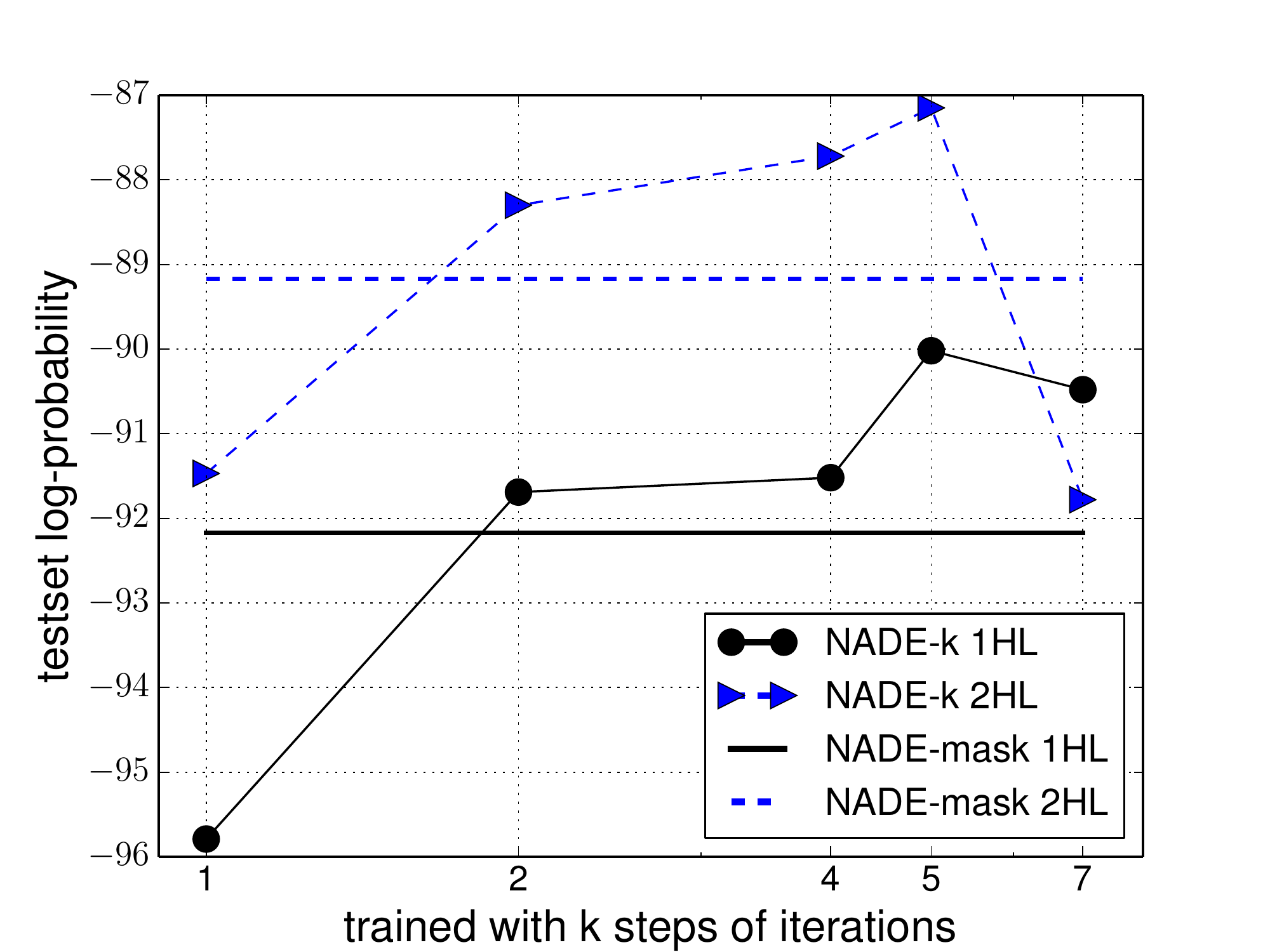}
(a)
\end{minipage}
\begin{minipage}{0.49\textwidth}
\centering
\includegraphics[width=\textwidth]{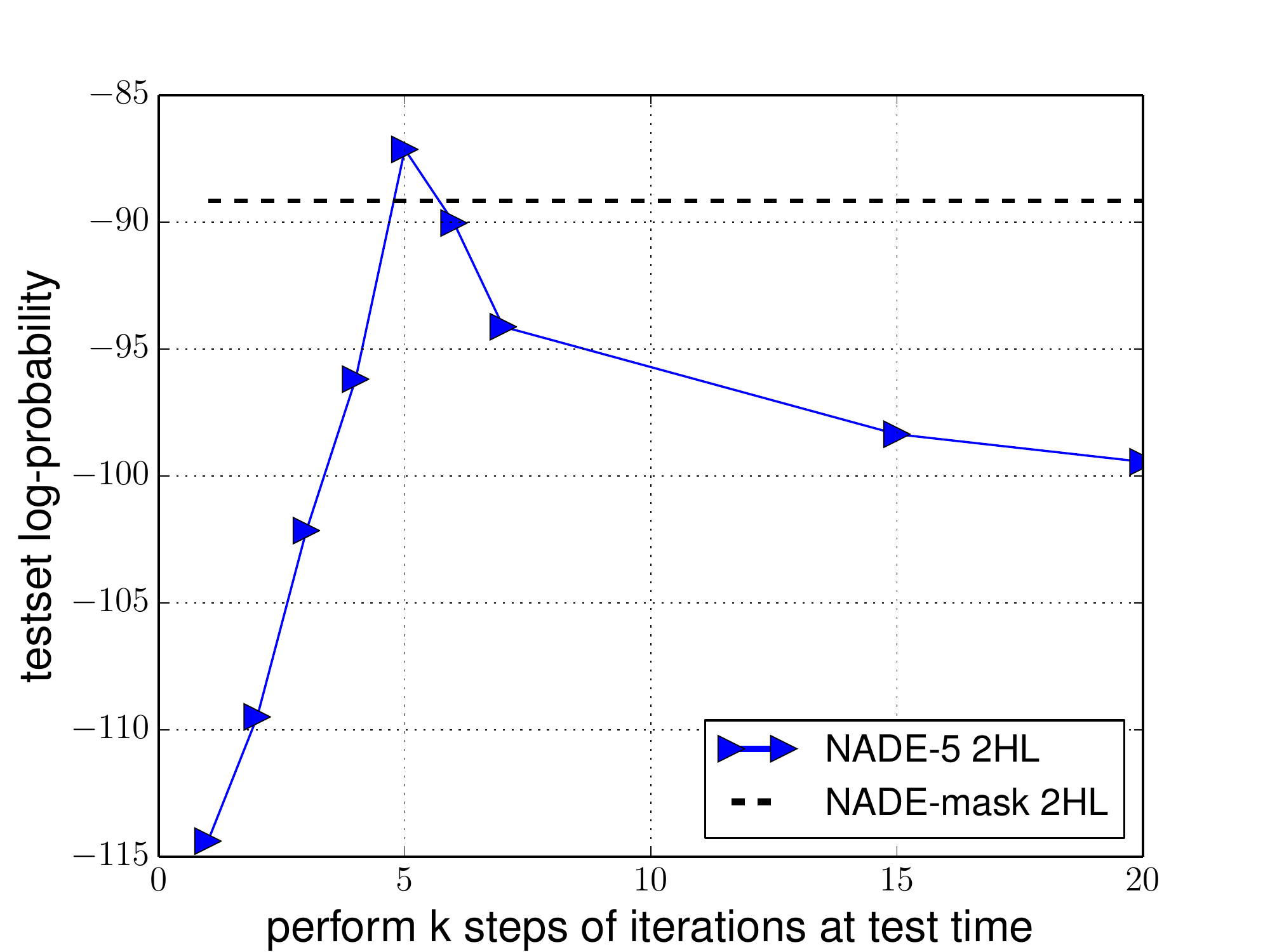} 
(b)
\end{minipage}
\caption{(a) The generalization performance of different \NADEk{} models trained with 
different $k$. (b) The generalization performance of NADE-5 2h, trained with k=5, but with 
various $k$ in test time.}
\label{fig:effect_of_k_in_train_test}
\end{figure}

Fig.~\ref{fig:effect_of_k_in_train_test} (a) shows the effect of the number of
iterations $k$ during training. Already with $k=2$, we can see that the \NADEk{}
outperforms its corresponding NADE-mask. The performance increases until $k=5$.
We believe the worse performance of $k=7$ is due to the well known training 
difficulty of a deep neural network,
considering that NADE-7 with two hidden layers effectively is a deep neural
network with $21$ layers.

At inference time, we found that it is important to use the exact $k$ that one
used to train the model. As can be seen from
Fig.~\ref{fig:effect_of_k_in_train_test} (b), the assigned probability increases
up to the $k$, but starts decreasing as the number of iterations goes over the
$k$. \footnote{In the future, one could explore possibilities for helping better
converge beyond step $k$, for instance by using costs based on reconstructions
at $k-1$ and $k$ even in the fine-tuning phase.}

\subsubsection{Qualitative Analysis}

In Fig.~\ref{fig:convergence}, we present how each iteration $t=1\dots k$
improves the corrupted input ($\vv^{\qt{t}}$ from Eq.~\eqref{eq:v0}).  We also
investigate what happens with test-time $k$ being larger than the training
$k=5$.  We can see that in all cases, the iteration -- which is a fixed
point update -- seems to converge to a point that is in most cases close to the
ground-truth sample.  Fig.~\ref{fig:effect_of_k_in_train_test} (b) shows however
that the generalization performance drops after $k=5$ when training with $k=5$.
From Fig.~\ref{fig:convergence}, we can see that the reconstruction continues to
be {\em sharper} even after $k=5$, which seems to be the underlying reason for
this phenomenon.

\begin{figure}[t]
\begin{minipage}{0.49\textwidth}
\centering
\includegraphics[width=\textwidth]{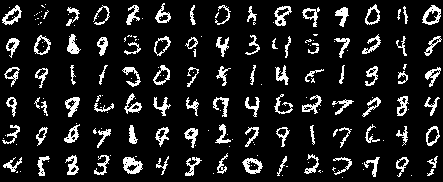}
(a) MNIST
\end{minipage}
\hfill
\begin{minipage}{0.49\textwidth}
\centering
\includegraphics[width=\textwidth]{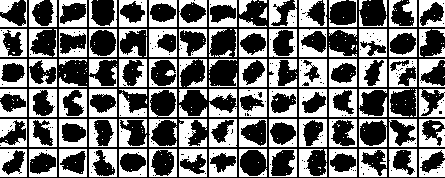} 
(b) Caltech-101 Silhouettes
\end{minipage}
\caption{Samples generated from \NADEk{} trained on (a) MNIST and (b) Caltech-101
Silhouettes.}
\label{fig:samples}
\end{figure}

\begin{figure}[t]
\begin{minipage}{0.49\textwidth}
\centering
\includegraphics[width=\textwidth]{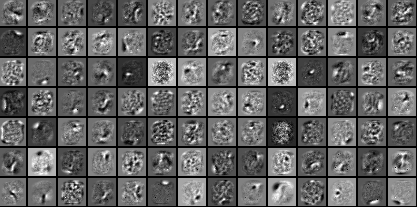} 
(a)
\end{minipage}
\begin{minipage}{0.49\textwidth}
\centering
\includegraphics[width=\textwidth]{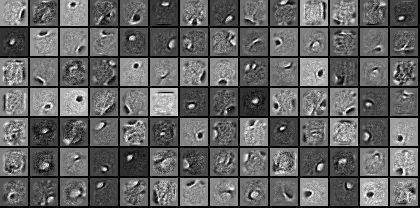}
(b)
\end{minipage}
\caption{Filters learned from NADE-5 2\hl. (a) A random subset of the encodering filters.
 (b) A random subset of the decoding filters.}
\label{fig:filters}
\end{figure}

From the samples generated from the trained NADE-$5$ with two hidden layers
shown in Fig.~\ref{fig:samples} (a), we
can see that the model is able to generate digits. Furthermore, the
filters learned by the model show that it has learned parts of digits such as
pen strokes (See Fig.~\ref{fig:filters}).

\subsubsection{Variability over Orderings}

In Section~\ref{seq:proposed_method}, we argued that we can perform any
inference task $p(\vx_\text{mis}\mid \vx_\text{obs})$ easily and efficiently by
restricting the set of orderings $O$ in Eq.~\eqref{eq:ensemble} to ones
where $\vx_\text{obs}$ is before $\vx_\text{mis}$.  For this to work well, we
should investigate how much the different orderings vary.

To measure the variability over orderings, we computed the variance of $\log
p(\vx \mid o)$ for 128 randomly chosen orderings $o$ with the trained \NADEk's
and NADE-mask with a single hidden layer. For comparison, we computed the
variance of $\log p(\vx \mid o)$ over the 10,000 test samples.

\begin{table}[ht]
    \centering
    \begin{minipage}{0.7\textwidth}
    \begin{tabular}{c | c c c }
        $\log p (\vx \mid o)$ & $\E_{o,\vx}\left[ \cdot \right]$ & $\sqrt{\E_\vx \var_{o}\left[ \cdot \right]}$ & $\sqrt{\E_o \var_\vx\left[ \cdot \right]}$ 
         \\
        \hline
        \hline
        NADE-mask 1\hl & -92.17 & 3.5 & 23.5   \\
        NADE-5 1\hl & -90.02 & 3.1 & 24.2  \\
        NADE-5 2\hl & -87.14 & 2.4 & 22.7  \\
    \end{tabular}
\end{minipage}
\hfill
\begin{minipage}{0.29\textwidth}
    \caption{The variance of $\log p(\vx \mid o)$ over orderings $o$ and over test samples $\vx$. }
    \label{tab:var_analysis}
\end{minipage}
\end{table}

In Table~\ref{tab:var_analysis}, the variability over the
orderings is clearly much smaller than that over the samples. Furthermore, the variability 
over orderings tends to decrease with the better models.

\subsection{Caltech-101 silhouettes}

We also evaluate the proposed \NADEk{} on Caltech-101
Silhouettes~\citep{Marlin10Inductive}, using the standard split of 4100 training
samples, 2264 validation samples and 2307 test samples. We demonstrate the advantage 
of \NADEk{} compared with NADE-mask under the constraint that they have a matching number of 
parameters. In particular, we compare \NADEk{} with 1000 hidden units with NADE-mask with 
670 
hiddens. We also compare \NADEk{} with 4000 hidden units with NADE-mask with 2670 hiddens.

We optimized the hyper-parameter 
$k \in \{1,2,\dots,10\}$ in the case of \NADEk{}. In both \NADEk{} and NADE-mask, 
we experimented without regularizations, with weight decays, or with dropout. 
Unlike the previous experiments, we
did not use the pretraining scheme (See Eq.~\eqref{eq:pretraining}). 

\begin{table}[ht]
\caption{Average log-probabilities of test samples of Caltech-101 Silhouettes.
    ($\star$) The results are from \citet{NECO_cho_2013_enhanced}. The terms 
in the parenthesis indicate the number of hidden units, the total number of parameters (M for million), and
the L2 regularization coefficient. NADE-mask 670h achieves the best performance without 
any regularizations.    
}
\hspace{2em} 
\centering 
\begin{tabularx}{\textwidth}{|X | c || X | c |}
\hline
Model  & Test LL & Model & Test LL\\
\hline
\hline
RBM$^\star$ \newline (2000h, 1.57M)       &  -108.98 & RBM $^\star$ \newline (4000h, 3.14M)       & {\bf -107.78} \\
\hline
NADE-mask \newline (670h, 1.58M)           & -112.51 & NADE-mask \newline (2670h, 6.28M, L2=0.00106)& -110.95\\
\cline{1-4}
NADE-2 \newline (1000h, 1.57M, L2=0.0054)  & -108.81 &
NADE-5 \newline (4000h, 6.28M, L2=0.0068)  & {\bf -107.28}  \\
\hline
\end{tabularx}
\label{tab:caltech_benchmark}
\end{table}   

As we can see from Table~\ref{tab:caltech_benchmark}, \NADEk{} 
outperforms the NADE-mask regardless of the number of parameters. In addition, 
NADE-2 with 1000 hidden units matches the performance of an RBM with the same number 
of parameters. Futhermore, NADE-5 has outperformed the previous best result 
obtained with the
RBMs in \citep{NECO_cho_2013_enhanced}, achieving the state-of-art result on this dataset. 
We can see from
the samples generated by the \NADEk{} shown in Fig.~\ref{fig:samples}
(b) that the model has learned the data well.

\section{Conclusions and Discussion}

In this paper, we proposed a model called iterative neural autoregressive
distribution estimator (\NADEk{}) that extends the conventional neural
autoregressive distribution estimator (NADE) and its order-agnostic training
procedure. The proposed \NADEk{} maintains the tractability of the original
NADE while we showed that it outperforms the original NADE as well as similar,
but intractable generative models such as restricted Boltzmann machines and
deep belief networks.

The proposed extension is inspired from the variational inference in
probabilistic models such as restricted Boltzmann machines (RBM) and deep
Boltzmann machines (DBM). Just like an iterative mean-field approximation in
Boltzmann machines, the proposed \NADEk{} performs multiple iterations through
hidden layers and a visible layer to infer the probability of the missing
value, unlike the original NADE which performs the inference of a missing value
in a single iteration through hidden layers.

Our empirical results show that this approach of multiple iterations improves
the performance of a model that has the same number of parameters, compared to
performing a single iteration.  This suggests that the inference method has
significant effect on the efficiency of utilizing the model parameters. Also,
we were able to observe that the generative performance of NADE can come close
to more sophisticated models such as deep belief networks in our approach.

In the future, more in-depth analysis of the proposed \NADEk{} is needed. For
instance, a relationship between \NADEk{} and the related models such as the
RBM need to be both theoretically and empirically studied.  The
computational speed of the method could be improved both in training (by using
better optimization algorithms. See, e.g.,~\citep{Pascanu+Bengio-ICLR2014}) and
in testing (e.g.\ by handling the components in chunks rather than fully
sequentially).  The computational efficiency of sampling for \NADEk{} can be 
further improved based on the recent work of \citet{yao2014equivalence} where 
an annealed Markov chain may be used to efficiently generate samples from the 
trained ensemble. Another promising idea to improve the model performance further is
to let the model adjust its own confidence based on $d$. For instance, in the
top right corner of Fig.\ \ref{fig:convergence}, we see a case with lots of
missing values values (low $d$), where the model is too confident about the
reconstructed digit 8 instead of the correct digit 2.

\subsubsection*{Acknowledgements}

The authors would like to acknowledge the support of NSERC, Calcul Qu\'{e}bec, Compute Canada, the Canada Research Chair and CIFAR, and developers of Theano \citep{bergstra+al:2010-scipy,Bastien-Theano-2012}. 









\newpage
\bibliography{strings,strings-shorter,ml,aigaion,read,myref}

\begin{thebibliography}{}

\bibitem[Bengio {\em et~al.}(2013)Bengio, Yao, Alain, and
  Vincent]{Bengio-et-al-NIPS2013-small}
Bengio, Y., Yao, L., Alain, G., and Vincent, P. (2013).
\newblock Generalized denoising auto-encoders as generative models.
\newblock In {\em Advances in Neural Information Processing Systems 26
  (NIPS'13)\/}.

\bibitem[Bengio {\em et~al.}(2014)Bengio, Thibodeau-Laufer, and
  Yosinski]{Bengio+Laufer+Yosinski-ICML-2014}
Bengio, Y., Thibodeau-Laufer, E., and Yosinski, J. (2014).
\newblock Deep generative stochastic networks trainable by backprop.
\newblock In {\em Proceedings of the 30th International Conference on Machine
  Learning (ICML'14)\/}.

\bibitem[Uria {\em et~al.}(2014)Uria, Murray, and
  Larochelle]{Benigno-et-al-icml-2014}
Uria, B., Murray, I., and Larochelle, H. (2014).
\newblock A deep and tractable density estimator.
\newblock In {\em Proceedings of the 30th International Conference on Machine
  Learning (ICML'14)\/}.

\end{thebibliography}


\begin{thebibliography}{}

\bibitem[Bastien {\em et~al.}(2012)Bastien, Lamblin, Pascanu, Bergstra,
  Goodfellow, Bergeron, Bouchard, and Bengio]{Bastien-Theano-2012}
Bastien, F., Lamblin, P., Pascanu, R., Bergstra, J., Goodfellow, I.~J.,
  Bergeron, A., Bouchard, N., and Bengio, Y. (2012).
\newblock Theano: new features and speed improvements.
\newblock Deep Learning and Unsupervised Feature Learning NIPS 2012 Workshop.

\bibitem[Bengio and Bengio(2000)Bengio and Bengio]{Bengio+Bengio-NIPS99}
Bengio, Y. and Bengio, S. (2000).
\newblock Modeling high-dimensional discrete data with multi-layer neural
  networks.
\newblock In {\em NIPS'99\/}, pages 400--406. MIT Press.

\bibitem[Bengio {\em et~al.}(2013)Bengio, Mesnil, Dauphin, and
  Rifai]{Bengio-et-al-ICML2013}
Bengio, Y., Mesnil, G., Dauphin, Y., and Rifai, S. (2013).
\newblock Better mixing via deep representations.
\newblock In {\em Proceedings of the 30th International Conference on Machine
  Learning (ICML'13)\/}. ACM.

\bibitem[Bergstra {\em et~al.}(2010)Bergstra, Breuleux, Bastien, Lamblin,
  Pascanu, Desjardins, Turian, Warde-Farley, and
  Bengio]{bergstra+al:2010-scipy}
Bergstra, J., Breuleux, O., Bastien, F., Lamblin, P., Pascanu, R., Desjardins,
  G., Turian, J., Warde-Farley, D., and Bengio, Y. (2010).
\newblock Theano: a {CPU} and {GPU} math expression compiler.
\newblock In {\em Proceedings of the Python for Scientific Computing Conference
  ({SciPy})\/}.
\newblock Oral Presentation.

\bibitem[Brakel {\em et~al.}(2013)Brakel, Stroobandt, and
  Schrauwen]{brakel2013training}
Brakel, P., Stroobandt, D., and Schrauwen, B. (2013).
\newblock Training energy-based models for time-series imputation.
\newblock {\em The Journal of Machine Learning Research\/}, {\bf 14}(1),
  2771--2797.

\bibitem[Cho {\em et~al.}(2013)Cho, Raiko, and Ilin]{NECO_cho_2013_enhanced}
Cho, K., Raiko, T., and Ilin, A. (2013).
\newblock Enhanced gradient for training restricted boltzmann machines.
\newblock {\em Neural computation\/}, {\bf 25}(3), 805--831.

\bibitem[Domke(2011)Domke]{domke2011parameter}
Domke, J. (2011).
\newblock Parameter learning with truncated message-passing.
\newblock In {\em Computer Vision and Pattern Recognition (CVPR), 2011 IEEE
  Conference on\/}, pages 2937--2943. IEEE.

\bibitem[Goodfellow {\em et~al.}(2013)Goodfellow, Mirza, Courville, and
  Bengio]{goodfellow2013multi}
Goodfellow, I., Mirza, M., Courville, A., and Bengio, Y. (2013).
\newblock Multi-prediction deep boltzmann machines.
\newblock In {\em Advances in Neural Information Processing Systems\/}, pages
  548--556.

\bibitem[Gregor {\em et~al.}(2014)Gregor, Danihelka, Mnih, Blundell, and
  Wierstra]{Gregor-et-al-ICML2014}
Gregor, K., Danihelka, I., Mnih, A., Blundell, C., and Wierstra, D. (2014).
\newblock Deep autoregressive networks.
\newblock In {\em International Conference on Machine Learning (ICML'2014)\/}.

\bibitem[Heckerman {\em et~al.}(2000)Heckerman, Chickering, Meek, Rounthwaite,
  and Kadie]{HeckermanD2000}
Heckerman, D., Chickering, D.~M., Meek, C., Rounthwaite, R., and Kadie, C.
  (2000).
\newblock Dependency networks for inference, collaborative filtering, and data
  visualization.
\newblock {\em Journal of Machine Learning Research\/}, {\bf 1}, 49--75.

\bibitem[Hinton(2000)Hinton]{Hinton-PoE-2000}
Hinton, G.~E. (2000).
\newblock Training products of experts by minimizing contrastive divergence.
\newblock Technical Report GCNU TR 2000-004, Gatsby Unit, University College
  London.

\bibitem[Huang and Ogata(2002)Huang and Ogata]{Huang2002}
Huang, F. and Ogata, Y. (2002).
\newblock Generalized pseudo-likelihood estimates for {M}arkov random fields on
  lattice.
\newblock {\em Annals of the Institute of Statistical Mathematics\/}, {\bf
  54}(1), 1--18.

\bibitem[Larochelle and Murray(2011)Larochelle and
  Murray]{larochelle2011neural}
Larochelle, H. and Murray, I. (2011).
\newblock The neural autoregressive distribution estimator.
\newblock {\em Journal of Machine Learning Research\/}, {\bf 15}, 29--37.

\bibitem[Marlin {\em et~al.}(2010)Marlin, Swersky, Chen, and
  de~Freitas]{Marlin10Inductive}
Marlin, B., Swersky, K., Chen, B., and de~Freitas, N. (2010).
\newblock Inductive principles for restricted {B}oltzmann machine learning.
\newblock In {\em Proceedings of The Thirteenth International Conference on
  Artificial Intelligence and Statistics (AISTATS'10)\/}, volume~9, pages
  509--516.

\bibitem[Pascanu and Bengio(2014)Pascanu and Bengio]{Pascanu+Bengio-ICLR2014}
Pascanu, R. and Bengio, Y. (2014).
\newblock Revisiting natural gradient for deep networks.
\newblock In {\em International Conference on Learning Representations
  2014(Conference Track)\/}.

\bibitem[Peterson and Anderson(1987)Peterson and Anderson]{Peterson1987}
Peterson, C. and Anderson, J.~R. (1987).
\newblock A mean field theory learning algorithm for neural networks.
\newblock {\em Complex Systems\/}, {\bf 1}(5), 995--1019.

\bibitem[Stoyanov {\em et~al.}(2011)Stoyanov, Ropson, and
  Eisner]{stoyanov2011empirical}
Stoyanov, V., Ropson, A., and Eisner, J. (2011).
\newblock Empirical risk minimization of graphical model parameters given
  approximate inference, decoding, and model structure.
\newblock In {\em International Conference on Artificial Intelligence and
  Statistics\/}, pages 725--733.

\bibitem[Uria {\em et~al.}(2014)Uria, Murray, and
  Larochelle]{Benigno-et-al-icml-2014}
Uria, B., Murray, I., and Larochelle, H. (2014).
\newblock A deep and tractable density estimator.
\newblock In {\em Proceedings of the 30th International Conference on Machine
  Learning (ICML'14)\/}.

\bibitem[Vincent {\em et~al.}(2010)Vincent, Larochelle, Lajoie, Bengio, and
  Manzagol]{Vincent-JMLR-2010-small}
Vincent, P., Larochelle, H., Lajoie, I., Bengio, Y., and Manzagol, P.-A.
  (2010).
\newblock Stacked denoising autoencoders: Learning useful representations in a
  deep network with a local denoising criterion.
\newblock {\em J. Machine Learning Res.}, {\bf 11}.

\bibitem[Yao {\em et~al.}(2014)Yao, Ozair, Cho, and Bengio]{yao2014equivalence}
Yao, L., Ozair, S., Cho, K., and Bengio, Y. (2014).
\newblock On the equivalence between deep nade and generative stochastic
  networks.
\newblock In {\em European Conference on Machine Learning (ECML/PKDD'14)\/}.
  Springer.

\bibitem[Zeiler(2012)Zeiler]{Zeiler-2012}
Zeiler, M.~D. (2012).
\newblock {ADADELTA}: an adaptive learning rate method.
\newblock Technical report, arXiv 1212.5701.

\end{thebibliography}
\bibliographystyle{natbib}

\end{document}